\documentclass[10pt,twocolumn,letterpaper]{article}

\usepackage{cvpr}              

\usepackage{graphicx}
\usepackage{amsmath}
\usepackage{amssymb}
\usepackage{booktabs}
\usepackage[table]{xcolor}
\usepackage{float}

\newcommand{\our}{{SG-GS}\xspace}

\definecolor{cvprblue}{rgb}{0.21,0.49,0.74}
\usepackage[pagebackref,breaklinks,colorlinks,allcolors=cvprblue]{hyperref}

\usepackage[capitalize]{cleveref}
\crefname{section}{Sec.}{Secs.}
\Crefname{section}{Section}{Sections}
\Crefname{table}{Table}{Tables}
\crefname{table}{Tab.}{Tabs.}

\def\cvprPaperID{2286} 
\def\confName{CVPR}
\def\confYear{2025}

\title{SG-GS: Topology-aware Human Avatars with Semantically-guided\\ Gaussian Splatting}

\author {
    Haoyu Zhao\textsuperscript{* \rm 1,\rm 2},
    Chen Yang\textsuperscript{* \rm 1},
    Hao Wang\textsuperscript{* \rm 3},
    Xingyue Zhao\textsuperscript{\rm 4},
    Wei Shen\textsuperscript{\textdagger \rm 1} \\
    \textsuperscript{1}MoE Key Lab of Artificial Intelligence, AI Institute, Shanghai Jiao Tong University \quad \\
    \textsuperscript{2}School of Computer Science, Wuhan University \quad \\
    \textsuperscript{3}Wuhan National Laboratory for Optoelectronics, Huazhong University of Science and Technology \quad \\
    \textsuperscript{4}School of Software Engineering, Xi’an Jiao Tong University \quad 
}

\begin{document}

\twocolumn[{%
    \renewcommand\twocolumn[1][]{#1}%
    \setlength{\tabcolsep}{0.0mm} 
    \newcommand{\sz}{0.125}  
    \maketitle
    \begin{center}
        \newcommand{\teaserwidth}{\textwidth}
        \includegraphics[width=\linewidth]{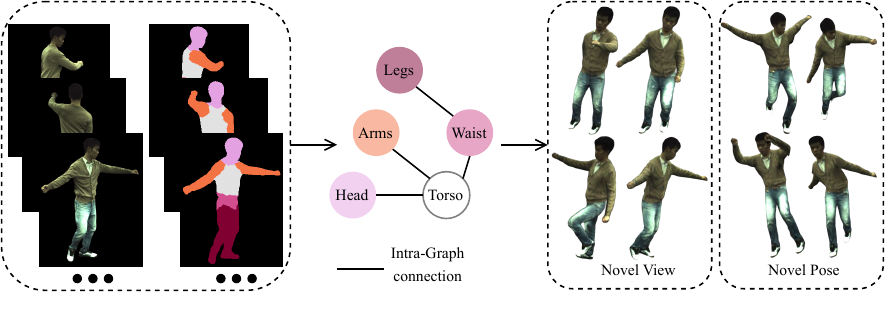}
    \captionof{figure}{We propose an efficient method for creating topology-aware human avatars from just videos, ensuring both photo-realistic human appearance and accurate anatomical structure. Our method achieve better quality to the most recent state-of-the-art methods~\cite{wen2024gomavatar,hu2024gauhuman,qian20243dgs}.} 
    \label{fig:first}
    \end{center}%
}]

\maketitle
{
\renewcommand{\thefootnote}{\fnsymbol{footnote}}
\footnotetext{* Equal contributions.}
\footnotetext{\textdagger Corresponding Author.}
\footnotetext{Haoyu Zhao completed this work during an internship at Shanghai Jiao Tong University.}
}
\maketitle

\begin{abstract}
Reconstructing photo-realistic and topology-aware animatable human avatars from monocular videos remains challenging in computer vision and graphics. 
Recently, methods using 3D Gaussians to represent the human body have emerged, offering faster optimization and real-time rendering. However, due to ignoring the crucial role of human body semantic information which represents the explicit topological and intrinsic structure within human body, they fail to achieve fine-detail reconstruction of human avatars. To address this issue, we propose \our, 
which uses semantics-embedded 3D Gaussians, skeleton-driven rigid deformation, and non-rigid cloth dynamics deformation to create photo-realistic human avatars. 
We then design a Semantic Human-Body Annotator (SHA) which utilizes SMPL's semantic prior for efficient body part semantic labeling. The generated labels are used to guide the optimization of semantic attributes of Gaussian. To capture the explicit topological structure of the human body, we employ a 3D network that integrates both topological and geometric associations for human avatar deformation. We further implement three key strategies to enhance the semantic accuracy of 3D Gaussians and rendering quality: semantic projection with 2D regularization, semantic-guided density regularization and semantic-aware regularization with neighborhood consistency.
Extensive experiments demonstrate that \our achieves state-of-the-art geometry and appearance reconstruction performance. Our project is at \href{https://sggs-projectpage.github.io/}{https://sggs-projectpage.github.io/}.
\end{abstract}

\section{Introduction}
Creating photo-realistic human avatars from monocular videos has immense potential value in industries such as gaming~\cite{zackariasson2012video}, extended reality storytelling~\cite{healey2021mixed}, and tele-presentation~\cite{ho2023learning}. In this work, we are dedicated to create high-quality photo-realistic human avatars from monocular videos with semantics embedded 3D Gaussians.

Recent advances in implicit neural fields~\cite{mildenhall2021nerf,wang2022r2l} enable high-quality reconstruction of geometry~\cite{xu2021h,wang2022arah,guo2023vid2avatar} and appearance~\cite{jiang2022neuman,li2022tava,yu2023monohuman,weng2022humannerf} of clothed human bodies from sparse multi-view or monocular videos. 
However, they often employ large MLPs, which makes training and rendering computationally demanding and inefficient.

Point-based rendering~\cite{zheng2023pointavatar} has emerged as an efficient alternative to NeRFs, offering significantly faster rendering speed. The recently proposed 3D Gaussian Splatting (3DGS)~\cite{kerbl20233d} achieves state-of-the-art novel view synthesis performance with significantly reduced inference time and faster training. 3DGS has inspired several recent works in human avatar creation~\cite{lei2024gart,moreau2024human,shao2024splattingavatar,hu2024gauhuman,kocabas2024hugs,wang2024gaussian,qian20243dgs,hu2024gaussianavatar}. However, these methods often overlook crucial semantic information that represents the explicit topological structure within the human body, leading to issues in maintaining anatomical coherence during motion and preserving fine details such as muscle definition and skin folds in various poses.

To this end, we propose \textbf{\our}, a \textbf{S}emantically-\textbf{G}uided 3D human model using \textbf{G}aussian \textbf{S}platting representation, as shown in Fig.~\ref{fig:first}. \our first integrates a skeleton-driven rigid deformation, and a non-rigid cloth dynamics deformation to coordinate the movements of individual Gaussians during animation. We then introduce a Semantic Human-Body Annotator (SHA), which leverages SMPL’s~\cite{loper2015smpl} human semantic prior for efficient body part semantic labeling. These part labels are used to guide the optimization of 3D Gaussian's semantic attribute. To learn topological relationships between human body parts, we propose a 3D topology- and geometry-aware network to learn body geometric and topological associations and integrate them into the avatar deformation. We further implement three key strategies to enhance semantic accuracy of 3D Gaussians and rendering quality: semantic projection with 2D regularization, semantic-guided density regularization and semantic-aware regularization with neighborhood consistency. 
Our experimental results demonstrate that \our achieves superior performance compared to current SOTA approaches in avatar creation from monocular inputs.
In summary, our work makes the following contributions:

\begin{itemize}
    \item We propose \our, which is the first to integrate semantic priors from the human body into creating animatable human avatars from monocular videos.
    \item We propose a 3D topology and geometry-aware network to capture topology and geometry information within the human body.
    \item We introduce semantic projection with 2D regularization, semantic neighborhood-consistent regularization, and semantic-guided density regularization to enhance semantic accuracy and rendering quality.
\end{itemize}


\section{Related Work}

\subsection{Neural Rendering for Human Avatars}
Since the introduction of Neural Radiance Fields (NeRF)~\cite{mildenhall2021nerf}, there has been a surge of research on neural rendering for human avatars~\cite{li2022tava,liu2021neural,li2023posevocab,peng2020neural,wang2022arah}. Though, NeRF is designed for static objects, HumanNeRF~\cite{weng2022humannerf} extend the NeRF to enable capturing a dynamic moving human using just a single monocular video. Neural Body~\cite{peng2020neural} associates a latent code to each SMPL~\cite{loper2015smpl} vertex to encode the appearance, which is transformed into observation space based on the human pose. Furthermore, Neural Actor~\cite{liu2021neural} learns a deformable radiance field with SMPL~\cite{loper2015smpl} as guidance and utilizes a texture map to improve its final rendering quality. Posevocab~\cite{li2023posevocab} designs joint-structured pose embeddings to encode dynamic appearances under different key poses, enabling more effective learning of joint-related appearances. However, a major limitation of NeRF-based methods is that NeRFs are slow to train and render.

Some works focus on achieving fast inference and training times for NeRF models of human avatars, including approaches that use explicit representations such as learning a function at grid points~\cite{chen2022tensorf}, using hash encoding~\cite{muller2022instant}, or altogether discarding the learnable component~\cite{fridovich2022plenoxels}. iNGP~\cite{muller2022instant} uses the underlying representation for articulated NeRFs, and enable interactive rendering speeds (15 FPS). \cite{chen2023uv} generates a pose-dependent UV volume, but its UV volume generation is not fast (20 FPS). In contrast to all these works, \our achieves state-of-the-art rendering quality and speed (25 FPS) with less training time.

\subsection{Dynamic 3D Gaussians for Human Avatars}
Point-based rendering~\cite{ruckert2022adop,zheng2023pointavatar} has proven to be an efficient alternative to NeRFs for fast inference and training. Extending point clouds to 3D Gaussians, 3D Gaussian Splatting (3DGS)~\cite{kerbl20233d} models the rendering process by splatting a set of 3D Gaussians onto the image plane via alpha blending. This approach achieves SOTA rendering quality with fast inference speed for novel views.

Given the impressive performance of 3DGS in both quality and speed, numerous works have further explored the 3D Gaussian representation for dynamic scene reconstruction. D-3DGS~\cite{luiten2023dynamic} is proposed as the first attempt to adapt 3DGS into a dynamic setup. Other works~\cite{wu20244d,yang2024deformable,zhao2024hfgs} model 3D Gaussian motions with a compact network or 4D primitives, resulting in highly efficient training and real-time rendering.

The application of 3DGS in dynamic 3D human avatar reconstruction is just beginning to unfold~\cite{jiang2024hifi4g,lei2024gart,qian20243dgs,hu2024gauhuman,kocabas2024hugs}. Human Gaussian Splatting~\cite{moreau2024human} showcase 3DGS as an efficient alternative to NeRF. Splattingavatar~\cite{shao2024splattingavatar} and Gomavatar~\cite{wen2024gomavatar} extends lifted optimization to simultaneously optimize the parameters of the Gaussians while walking on the triangle mesh. While these methods have made significant progress, they often overlook the crucial role of semantic information which is related to topological relationships between human body parts. It is a key focus of our \our.


\section{Preliminaries}

\medskip
\noindent
\textbf{SMPL~\cite{loper2015smpl}.} The SMPL model is a widely-used parametric 3D human body model that efficiently represents body shape and pose variations. In our work, We utilize SMPL's Linear Blend Skinning (LBS) algorithm to transform points from canonical space to observation space, enabling accurate body deformation across different poses. We also leverage SMPL's body priors to enhance the model’s understanding of body structure, improving the quality and consistency of human avatar reconstruction.

\medskip
\noindent
\textbf{3D Gaussian Splatting (3DGS)~\cite{kerbl20233d}.} 3DGS explicitly represents scenes using point clouds, where each point is modeled as a 3D Gaussian defined by a covariance matrix $\Sigma$ and a center point $\mathcal{X}$, the latter referred to as the mean. The value at point $\mathcal{X}$ is: $G(\mathcal{X})=e^{-\frac{1}{2}\mathcal{X}^T\Sigma^{-1}\mathcal{X}}$.

For differentiable optimization, the covariance matrix $\Sigma$ is decomposed into a scaling matrix $\mathcal{S}$ and a rotation matrix $\mathcal{R}$, such that $\Sigma = \mathcal{R}\mathcal{S}\mathcal{S}^T\mathcal{R}^T$.
In practice, $\mathcal{S}$ and $\mathcal{R}$ are also represented by the diagonal vector $s\in \mathbb{R}^{N\times3}$ and a quaternion vector $r\in \mathbb{R}^{N\times4}$, respectively.
In rendering novel views, differential splatting, as introduced by~\cite{yifan2019differentiable} and~\cite{zwicker2001surface}, involves applying a viewing transformation $W$ along with the Jacobian matrix $J$ of the affine approximation of the projective transformation. This process computes the transformed covariance matrix as: $\Sigma^{\prime} = JW\Sigma W^TJ^T$.
The color and opacity at each pixel are computed from the Gaussian’s representation $G(\mathcal{X})=e^{-\frac{1}{2}\mathcal{X}^T\Sigma^{-1}\mathcal{X}}$. The pixel color $\mathcal{C}$ is computed by blending $N$ ordered 3D Gaussian splats that overlap at the given pixel, using the formula:
\begin{equation}
    \mathcal{C} = \sum_{i\in N}c_i \alpha_i \prod_{j=1}^{i-1} (1-\alpha_i).
\label{eq:gaussian_render}
\end{equation}
Here, $c_i$, $\alpha_i$ represents the density and color of this point computed by a 3D Gaussian $G$ with covariance $\Sigma$ multiplied by an optimizable per-point opacity and SH color coefficients. The 3D Gaussians are optimized using a photometric loss. 3DGS adjusts their number through periodic densification and pruning, achieving an optimal density distribution that accurately represents the scene.


\begin{figure*}[ht]
  \centering
    \includegraphics[width=\linewidth]{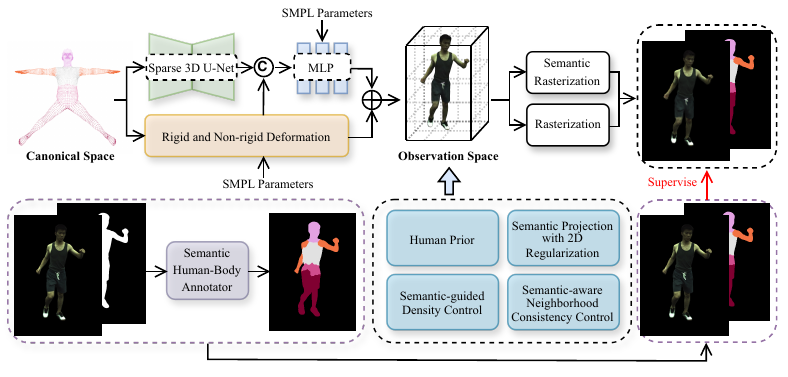}\\
  \caption{\textbf{Our framework for creating photo-realistic animatable avatars from monocular videos.} We initialize a set of 3D Gaussians in the canonical space by sampling 6,890 points from the SMPL model and assign the semantic attributes of Gaussians to each point. We first integrate a skeleton-driven rigid deformation and a non-rigid cloth dynamics deformation to deform human avatars from canonical space $\mathcal{G}_c$ to observation space $\mathcal{G}_o$. Then, we introduce a Semantic Human-Body Annotator (SHA), which leverages SMPL’s human body semantic prior for efficient semantic labeling. These labels are used to guide the optimization of 3D Gaussian’s semantic attribute $\mathcal{O}$. We also propose a 3D topology and geometry-aware network to learn body topological and geometric associations and integrate them into learning the 3D deformation. To enhance semantic accuracy and render quality, we implement semantic projection with 2D regularization, semantic-guided density regularization and semantic-aware regularization with neighborhood consistency.}
  \label{fig:pipeline}
\end{figure*}




\section{Method}
In this section, we illustrate the pipeline of our \our in Fig.~\ref{fig:pipeline}. The inputs to our method include images $X = \{x_i\}^N_{i=1}$ obtained from monocular videos, fitted SMPL parameters $P = \{p_i\}^N_{i=1}$, and paired foreground masks $M = \{m_i\}^N_{i=1}$ of images. \our optimizes 3D Gaussians in canonical space, which are then deformed to match the observation space and rendered from the provided camera view. For a set of 3D Gaussians, we store the following properties at each point: position $\mathcal{X} \in \mathbb{R}^3$, color defined by spherical harmonic (SH) coefficients $\mathcal{C} \in \mathbb{R}^k$ (where $k$ is the number of SH functions), opacity $\alpha \in \mathbb{R}$, rotation factor $r \in \mathbb{R}^4$, and scaling factor $s \in \mathbb{R}^3$. To integrate semantic information about body parts into the 3D Gaussian optimization process and learn the topological structure of the human body, we divide the human body into 5 distinct parts, as shown in Fig.~\ref{fig:first}. We represent the labels using one-hot encoding, stored as semantic attribute $\mathcal{O} \in \mathbb{R}^{10}$.



\subsection{Non-rigid and Rigid Deformation}
\label{sec: deformation}

Inspired by~\cite{weng2022humannerf,qian20243dgs}, We decompose human deformation into two key components: 1) a non-rigid element capturing pose-dependent cloth dynamics, and 2) a rigid transformation governed by the human skeletal structure. 

We employ a non-rigid deformation network, that takes the canonical position $\mathcal{X}_c$ of the 3D Gaussians ${\mathcal{G}_c}$ in canonical space and a pose latent code as input. This pose latent code encodes SMPL parameters $p_i$ using a lightweight hierarchical pose encoder~\cite{mihajlovic2021leap} into $\mathcal{Z}_p$. The network then outputs offsets for various parameters of ${\mathcal{G}_c}$: 
\begin{align}
\Delta {(\mathcal{X},\mathcal{C},\alpha,s,r)} = f_{\theta_{nr}}\left(\mathcal{X}_c;\mathcal{Z}_p\right).
\label{eq:mlp}
\end{align}
This network enables efficient and detailed non-rigid deformation of the 3D Gaussians, effectively capturing the nuances of human body movement and shape. The canonical Gaussian is deformed by:
\begin{align}
\mathcal{X}_d &= \mathcal{X}_c + \Delta \mathcal{X}, \mathcal{C}_d = \mathcal{C}_c + \Delta \mathcal{C} \label{eq:deform1}, \\
\alpha_d &= \alpha_c + \Delta \alpha, s_d = s_c + \Delta s, \label{eq:deform2} \\
r_d &= r_c \cdot [1,\Delta r_1,\Delta r_2, \Delta r_3], \label{eq:deform3}
\end{align}
where the quaternion multiplication $\cdot$ is equivalent to multiplying the corresponding rotation matrices. With $[1,0,0,0]$ representing the identity rotation, $r_d=r_c$ when $\delta{r}=\mathbf{0}$, preserving the original orientation for zero rotation offset.

We further employ a rigid deformation network to transform the non-rigidly deformed 3D Gaussians ${\mathcal{G}_d}$ to the observation space ${\mathcal{G}_o}$. This is achieved via forward Linear Blend Skinning (LBS):
\begin{align}
    &\mathbf{T} = \sum_{b=1}^B f_{\theta_r}(\mathcal{X}_d)_b \mathbf{B}_b,  
    \mathcal{X}_o = \mathbf{T}\mathcal{X}_d \label{eq:r_gaussian_xyz}, \\
    &\mathcal{R}_o = \mathbf{T}_{1:3,1:3}\mathcal{R}_d \label{eq:r_gaussian_rot},
\end{align}
where $\mathcal{R}_d$ is the rotation matrix derived from quaternion $r_d$, and $\mathbf{B}_b$ represents the differentiable bone transformations. This step aligns the deformed Gaussians with the target pose in the observation space ${\mathcal{G}_o}$.

\subsection{Semantic Human-Body Annotator}
\label{sec:sha}
Most current animatable human avatar creation methods just use SMPL~\cite{loper2015smpl} model for its pose-aware shape priors, neglecting its inherent semantic information. We argue that semantic information contains topological relationships within human body which can improve rendering quality during complex motion deformations. We will further demonstrate this in the experimental Section.~\ref{sec:ablation}. 

To achieve this, we deform the standard human body model from the SMPL model using the differentiable bone transformations $\mathbf{B}_b$ as described in Eq.~\ref{eq:r_gaussian_xyz}. Then, we use a custom point rasterizing function to render the deformed 3D SMPL model into an image $m^p_i$ with a projection matrix from the dataset.

For each pixel in a foreground mask $m_i$, we employ the k-nearest neighbors (KNN) algorithm to identify the closest pixels in $m^p_i$. This process enables semantic-level annotation of body parts by transferring semantic labels from the SMPL model to the foreground mask $m_i$. The result is a semantically annotated mask $m^s_i$ that accurately represents the different regions of the human body. We formalize this Semantic Human Annotation (SHA) process as follows:
\begin{equation}
    m^s_i = \mathcal{SHA}(m_i, \mathbf{B}_b),
    \label{eq:semantic_annotation}
\end{equation}
where $\mathcal{SHA}$ denotes our Semantic Human-Body Annotator. We use the generated human body semantic labels $m_s$ to supervise the Gaussian's semantic attribute. (described in semantic projection with 2D regularization in Section.~\ref{sec:optimization}).



While there are pre-trained networks for human parsing, such as SCHP~\cite{li2020self} and Graphonomy~\cite{gong2019graphonomy}, they are designed to segment both clothing and human body parts jointly. In contrast, our work focuses on leveraging semantic information to learn the topological relationships between different body parts. The objectives and tasks of these networks do not fully align with our needs, limiting their ability to model the geometric structure and topological connections of the human body. Their clothing segmentation can also introduce noise, hindering accurate body topology learning.


\subsection{Topological and Geometric Feature Learning}
\label{sec:geometric-semantic}

To jointly learn and embed topology and geometry information to human avatar deformation, we propose a 3D topology- and geometry-aware network that effectively captures the human body's local topological and geometric structure in canonical space.

We treat 3D Gaussians as a point cloud. Point-level MLPs are limited by a small receptive field, which restricts their capability to capture the local geometric and topological features. Therefore, we employ sparse convolution~\cite{graham2017submanifold} on sparse voxels to extract local topological and geometric features across varying receptive fields, following the method outlined in~\cite{lu20243d}. Given the position $\mathcal{X}_c$ of the Gaussians $\mathcal{G}_c$ as a point cloud, we initially convert it into voxels by partitioning the space using a fixed grid size $v$.
\begin{equation}
    \mathbf{V} = \lfloor \mathcal{X}_c / v \rfloor,
    \label{eq:voxelization}
\end{equation}
where $\mathbf{V} \in \mathbb{R}^{M \times 3}$ and $M$ is the number of voxels. We then construct a 3D sparse U-Net by stacking a series of sparse convolutions with skip connections to aggregate local features. 
The sparse 3D U-Net $f_{\theta_{unet}}$ takes $\mathbf{V}$ and the semantic point-based features $\mathcal{O}$ as input, and outputs topological and geometric features $\mathbf{F}_v$:
\begin{equation}
    \mathbf{F}_v = f_{\theta_{unet}}(\mathbf{V}; \mathcal{O}).
\end{equation}
We process the feature $\mathbf{F}_v$, the position $\mathcal{X}_d$ of the deformed Gaussians $\mathcal{G}_o$, and pose latent code $\mathcal{Z}_p$ in Eq.~\ref{eq:mlp} through an fusion network $f_{\theta_{sr}}$:
\begin{equation}
    \Delta {(\mathcal{X}',s',r')} = f_{\theta_{sr}}(\mathbf{F}_v; \mathcal{X}_d;\mathcal{Z}_p),
    \label{eq:feature_fusion}
\end{equation}
where $\Delta {(\mathcal{X}',s',r')}$ represents the final fused features. The deformed 3D Gaussian $\mathcal{G}_o$ is then deformed by $\Delta {(\mathcal{X}',s',r')}$ following Eq.~\ref{eq:deform1}, \ref{eq:deform2}, and~\ref{eq:deform3}.


\subsection{Optimization}
\label{sec:optimization}


Unlike random initialization or Structure-from-Motion (SfM) initialization for Gaussian point clouds, we directly sample 6,890 points from the SMPL model~\cite{loper2015smpl} as our initial point cloud. Each Gaussian is then assigned semantic attributes based on the SMPL model's predefined semantic labels.During densification, newly created 3D Gaussian points inherit semantic attributes from their parent nodes.



\medskip
\noindent
\textbf{Semantic projection with 2D regularization.} 
We acquire rendered per-pixel semantic labels using the efficient Gaussian splatting algorithm following Eq.~\ref{eq:gaussian_render} as:
\begin{equation}
\mathcal{S} = \sum_{g \in \mathcal{N}} \mathcal{O}_{g} \alpha_{g} \prod_{j=1}^{g-1}\left(1-\alpha_{j}\right),
\end{equation}
where $\mathcal{S}_k$ represents the 2D semantic labels of pixel $k$, derived from Gaussian point semantic attributes via $\alpha$-blending (Eq.~\ref{eq:gaussian_render}). Here, $\mathcal{O}_g$ denotes the semantic attribute of the 3D Gaussian point $g$, and $\alpha_g$ is the influence factor of this point in rendering pixels. Upon calculating these labels, we obtain the results $l^s_i$ and apply a BCE loss to regularize the rendered semantic label $l^s_i$ with semantic labels generated via SHA as follows:
\begin{equation}
    \mathcal{L}_{semantic} = \mathcal{L}_{bce}(l^s_i, m^s_i).
\label{eq:semantic}
\end{equation}


\medskip
\noindent
\textbf{Semantic-guided density regularization.} 
Fuzzy geometric shapes often appear in local structures on the human surface, particularly in high-frequency areas like clothing wrinkles and muscle textures~\cite{wang2024gaussian}. To improve the clarity and distribution of 3D Gaussians in these regions, we propose semantic-guided density regularization. We identify high-frequency nodes by assessing the average magnitude of structural differences between a selected node and all nodes within the same cluster. Nodes exhibiting the highest average magnitude of these differences are designated as high-frequency nodes.
\begin{equation}
    H_m = \underset{i \in C_m}{\operatorname{arg\,max}} \left\{ \frac{1}{|C_m| - 1} \sum_{j \in C_m \setminus \{i\}} d(A_i, A_j) \right\},
    \label{eq:high_freq_node}
\end{equation}

\noindent where $H_m$ is the high-frequency node in cluster $C_m$, $A_i$ is the basic attribute of 3D Gaussian points (color, opacity, etc.), $C_m$ represents the set of all points with semantic attribute $m$, $C_m \setminus \{i\}$  denotes the set of elements in  $C_m$  excluding the element  $i$, and $d(\cdot, \cdot)$ is a dissimilarity measure between two points. To better capture and express these local structures of significant discrepancies, we perform densification operations on these 3D Gaussians, enhancing the local rendering granularity to focus on guiding the split and attribute optimization of Gaussian points in these areas.


\medskip
\noindent
\textbf{Semantic-aware regularization with neighborhood consistency.} 
We expect Gaussians that are in close proximity to exhibit similar semantic attributes, thereby achieving local semantic consistency in 3D space. The loss function for this semantic consistency constraint is as follows:
\begin{equation}
\mathcal{L}_{neighborhood} = \frac{1}{|N|} \sum_{m \in N} \sum_{n \in N_k(m)} D_{\text{KL}}(\mathcal{O}_m || \mathcal{O}_n),
\label{eq:simplified_neighborhood_consistency}
\end{equation}
where $N$ represents the total number of Gaussian points, $N_k(m)$ contains the $k$ nearest neighbors of 3D Gaussian point $m$ in 3D space, $\mathcal{O}_m$ and $\mathcal{O}_n$ represent the predicted semantic attribute for point $m$ and its neighbor $n$, respectively, and $D_{\text{KL}}(q_m || q_n)$ calculates the KL divergence between the predicted distributions of point $m$ and its neighbor $n$.




\begin{figure*}[!htb]
  \centering
    \includegraphics[width=\linewidth]{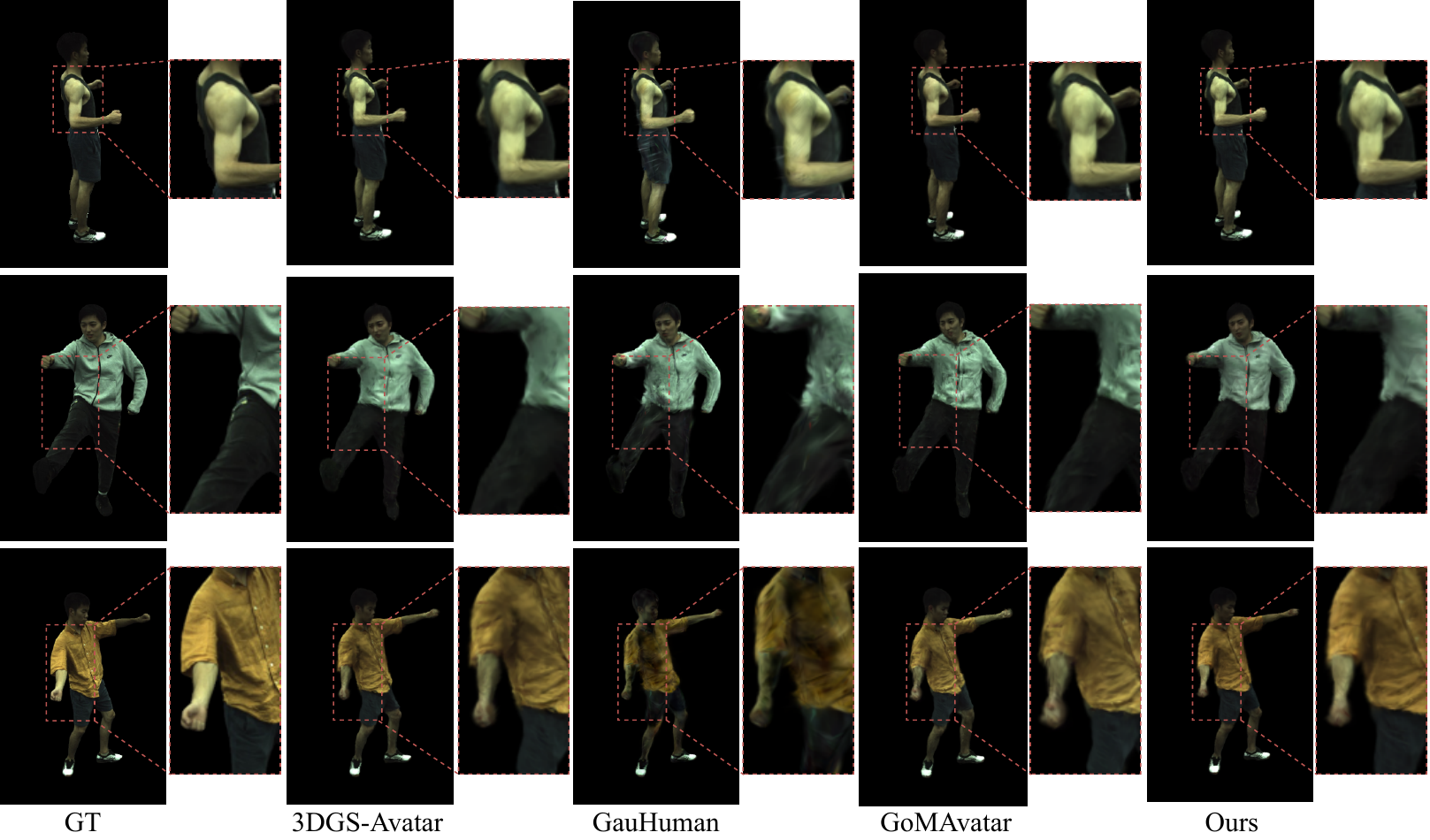}\\
  \caption{\textbf{Qualitative Comparison on ZJU-MoCap~\cite{peng2020neural}.} We show that our \our can produce realistic details in both rendered images and geometry, while other approaches struggle to generate smooth details.}
  \label{fig:result}
\end{figure*}

\medskip
\noindent
\textbf{Loss function.} 
Our full loss function consists of a RGB loss $\mathcal{L}_{rgb}$, a mask loss $\mathcal{L}_{mask}$, a skinning weight regularization loss $\mathcal{L}_{skin}$, the as-isometric-as-possible regularization loss $\mathcal{L}_{isopos}$ following~\cite{qian20243dgs}, Semantic projection with 2D regularization $\mathcal{L}_{semantic}$, and Semantic-aware regularization with neighborhood consistency $\mathcal{L}_{neighborhood}$:
\begin{align}
\mathcal{L}_{reconstruct} =  &\mathcal{L}_{rgb} + \lambda_1 \mathcal{L}_{mask} + \lambda_2 \mathcal{L}_{SSIM} + \lambda_3 \mathcal{L}_{LPIPS} \nonumber\\
 & + \lambda_4 \mathcal{L}_{skin} + \lambda_5 \mathcal{L}_{isopos}.
\label{eq:loss}
\end{align}
The final loss function is:
\begin{align}
\mathcal{L} = \mathcal{L}_{reconstruct}+\lambda_6 \mathcal{L}_{semantic} +\lambda_7 \mathcal{L}_{neighborhood},
\end{align}
where $\lambda$'s are loss weights. For further details of the loss definition and respective weights, please refer to the Supp.Mat.




\begin{table}[t]
\setlength{\fboxsep}{2pt}
\fontsize{9}{10}\selectfont
 \caption{\textbf{Quantitative Results on ZJU-MoCap~\cite{peng2020neural}.} \our achieves state-of-the-art performance across every method. The \colorbox{pink}{best} and the \colorbox{yellow}{second best} results are denoted by pink and yellow. Frames per second (FPS) is measured on an RTX 3090. LPIPS* = LPIPS $\times$ 1000.}
 \label{tab:compare_zjumocap}
 \centering
 \setlength{\tabcolsep}{2pt}
 \renewcommand{\arraystretch}{1.1}
 \begin{tabular}{ l|cccc}
 \toprule
Method:                                 
& PSNR$\uparrow$
& SSIM$\uparrow$
& LPIPS*$\downarrow$    
& FPS 
 \\\hline
 NeuralBody~\cite{peng2020neural}  & 29.07 & 0.962 & 52.29  & 1.5\\ 
 Ani-NeRF~\cite{peng2021animatable} & 29.17 & 0.961 & 51.98 & 1.1\\
 HumanNeRF~\cite{weng2022humannerf} & 30.24 & \cellcolor{yellow}0.968 & 31.73 & 0.3\\
 MonoHuman~\cite{yu2023monohuman} & 29.38 & 0.964 & 37.51 & 0.1 \\
  DVA~\cite{remelli2022drivable} & 29.45 & 0.956 & 37.74 & 17 \\
  InstantAvatar~\cite{jiang2022instantavatar} & 29.73 & 0.938 & 64.41 & 4.2\\
 3DGS-Avatar~\cite{qian20243dgs}   & 30.62 & 0.965 & \cellcolor{yellow}30.28 & \cellcolor{yellow}50\\
 GauHuman~\cite{hu2024gauhuman} & \cellcolor{yellow}30.79 & 0.960 & 32.73 & \cellcolor{pink}180\\ 
 GoMAvatar~\cite{wen2024gomavatar} & 30.37 & \cellcolor{pink}0.969 & 32.53 & 43 \\
 \our & \cellcolor{pink}30.88 & \cellcolor{pink}0.969 & \cellcolor{pink}29.69 & 25\\
 \bottomrule
 \end{tabular}
\end{table}



\begin{table}
\setlength{\fboxsep}{2pt}
\fontsize{9}{10}\selectfont
 \caption{\textbf{Quantitative Results on H36M~\cite{ionescu2013human3}.} Our \our still achieves superior performance compared to state-of-the-art methods on both training poses and novel poses.}
 \label{tab:compare_h36m}
 \centering
 \setlength{\tabcolsep}{1pt}
 \renewcommand{\arraystretch}{1.1}
 \begin{tabular}{ l|cc|cc}
 \toprule                  
 & \multicolumn{2}{c}{Training Poses}                                 
 & \multicolumn{2}{c}{Novel Poses}                                 
 \\ 
Method:             
& PSNR$\uparrow$
& SSIM$\uparrow$   
& PSNR$\uparrow$
& SSIM$\uparrow$  
 \\\hline
NARF~\cite{noguchi2021neural} & 23.00 & 0.898 & 22.27 & 0.881 \\
NeuralBody~\cite{peng2020neural} & 22.89 & 0.896 & 23.09 & 0.891 \\
Ani-NeRF~\cite{peng2021animatable} & 23.00 & 0.890 & 22.55 & 0.880 \\
ARAH~\cite{ARAH:ECCV:2022} & 24.79 & 0.918 & 23.42 & 0.896  \\
3DGS-Avatar~\cite{qian20243dgs} & \cellcolor{yellow}32.89 & \cellcolor{yellow}0.982 & \cellcolor{yellow}32.50 & \cellcolor{yellow}0.983  \\
 \our & \cellcolor{pink}33.01 & \cellcolor{pink}0.989 & \cellcolor{pink}33.14 & \cellcolor{pink}0.987     \\
 \bottomrule
  \end{tabular}
\end{table}


\begin{figure}[t]
  \centering
    \includegraphics[width=\linewidth]{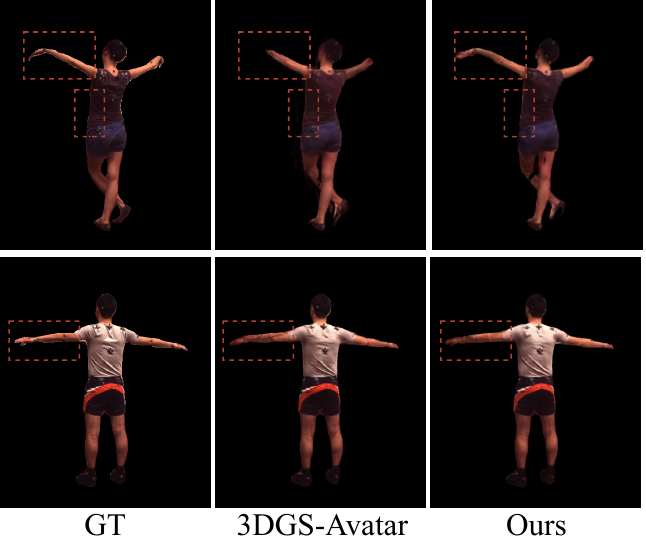}\\
  \caption{\textbf{Qualitative Comparison on H36M~\cite{ionescu2013human3}.} By utilizing semantic information within human body, our \our preserves better anatomical structures of the human body, producing high-quality results}
  \label{fig:result_h36m}
\end{figure}

\begin{table}[t]
 \caption{\textbf{Ablation Study on ZJU-MoCap~\cite{peng2020neural}.} 
 The proposed model achieves the lowest LPIPS, demonstrating the effectiveness of all components.}
 \label{tab:ablation}
 \centering
\setlength{\fboxsep}{3pt}
\fontsize{9}{10}\selectfont
 \begin{tabular}{@{}lcccc}
 \toprule
 Method:          
 & PSNR$\uparrow$
 & SSIM$\uparrow$
 & LPIPS*$\downarrow$
 & FPS \\
 \hline
 Baseline & 30.17 & 0.961 & 36.57 & 70\\
 w/o topo-geo & 30.64 & \cellcolor{pink}0.970 & 32.75 & 70\\
 mlp & 30.54 & 0.967 & 32.06 & 60\\
 w/o $\mathcal{L}_{semantic}$ & \cellcolor{yellow}30.67 & 0.966 & \cellcolor{yellow}29.99 & 25\\
 w/o sgd & 30.55 & 0.968 & 31.01 & 25\\
 w/o $\mathcal{L}_{neighborhood}$ & 30.56 & 0.965 & 30.59 & 25\\
 \our & \cellcolor{pink}30.88 & \cellcolor{yellow}0.969 & \cellcolor{pink}29.69 & 25\\
 \bottomrule
 \end{tabular}
\end{table}

\section{Experiment}

In this section, we first compare \our with recent SOTA methods~\cite{peng2020neural,peng2021animatable,weng2022humannerf,ARAH:ECCV:2022,yu2023monohuman,qian20243dgs,hu2024gauhuman,hu2024gaussianavatar}, demonstrating that our \our achieves superior rendering quality. We then systematically ablate each component of the proposed method, showing their effectiveness in better rendering performance. All models are trained on one single NVIDIA RTX 3090 GPU. For further details of implementation, please refer to the Supp.Mat. 


\subsection{Dataset}

\medskip
\noindent
\textbf{ZJU-MoCap~\cite{peng2020neural}.} 
It records multi-view videos with 21 cameras and collects human poses using the marker-less motion capture system. We select six sequences (377, 386, 387, 392, 393, 394) from this dataset to conduct experiments. We also follow the same training/test split following~\cite{weng2022humannerf,qian20243dgs}, \textit{i.e.}, one camera is used for training, while the remaining cameras are used for evaluation.

\medskip
\noindent
\textbf{H36M~\cite{ionescu2013human3}.}
It captures multi-view videos using four cameras and collects human poses with a marker-based motion capture system. It includes multiple subjects performing complex actions. We select representative actions, split the videos into training and test frames, following ARAH~\cite{ARAH:ECCV:2022}, and perform experiments on sequences (S1, S5, S6, S7, S8, S9, S11). Three cameras are used for training and the remaining is selected for test. 



\subsection{Comparison with State-of-the-art Methods}
We conduct comparative experiments against various state-of-the-art (SOTA) methods for human avatars, including NeRF-based methods such as NeuralBody~\cite{peng2020neural}, Ani-NeRF~\cite{peng2021animatable}, HumanNeRF~\cite{weng2022humannerf}, and MonoHuman~\cite{yu2023monohuman}, as well as 3DGS-based methods such as 3DGS-Avatar~\cite{qian20243dgs}, GauHuman~\cite{hu2024gauhuman}, and GoMAvatar~\cite{wen2024gomavatar}, under a monocular setup on ZJU-MoCap~\cite{peng2020neural}. In Table~\ref{tab:compare_zjumocap}, we evaluate the reconstruction quality using three different metrics: PSNR, SSIM, and LPIPS. Thanks to the LBS weight field and deformation field learned in HumanNeRF\cite{weng2022humannerf}, 3DGS-Avatar~\cite{qian20243dgs}, and GauHuman~\cite{hu2024gauhuman}, these methods achieve comparable visualization results. In comparison, our proposed \our achieves good performance in terms of PSNR and SSIM while significantly outperforming existing methods on LPIPS. Existing researches~\cite{yang2024gaussianobject,qian20243dgs} reach a consensus that \textit{LPIPS provides more meaningful insights compared to the other metrics, given the challenges of reproducing exact ground-truth appearances for novel views.} 

As shown in Fig.~\ref{fig:result}, our \our method preserves sharper details compared to other methods. \textit{Notably, our approach excels at capturing fine details in challenging areas such as clothing, where reconstruction is typically more difficult due to intricate textures.} By preserving these finer details, our method provides a more realistic and detailed reconstruction of clothing and other complex surfaces, significantly improving the overall quality and fidelity of the 3D human avatars. Please see our project website videos and supplementary material for more video visualization.

In addition, we also evaluate our \our using the H36M~\cite{ionescu2013human3} dataset. We report the quantitative results against NeRF-based methods such as NARF~\cite{noguchi2021neural}, NeuralBody~\cite{peng2020neural}, Ani-NeRF~\cite{peng2021animatable}, and ARAH~\cite{ARAH:ECCV:2022}, as well as 3DGS-based methods such as 3DGS-Avatar~\cite{qian20243dgs} in Table~\ref{tab:compare_h36m}. Our model outperforms both established NeRF-based methods and 3DGS-based methods. As shown in Fig.~\ref{fig:result_h36m}, due to the use of semantic information within human body, our \our achieves better reconstruction of edge areas and preserves anatomical structures of the human body.


\begin{figure}[t]
  \centering
    \includegraphics[width=\linewidth]{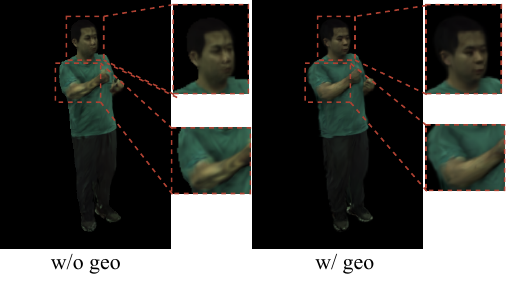}\\
  \caption{\textbf{Ablation Study} on Geometric and Semantic Feature Learning, which helps erase artifacts and learn fine details like cloth wrinkles and human face under novel views.}
  \label{fig:geo}
\end{figure}


\begin{figure}[ht]
  \centering
    \includegraphics[width=0.75\linewidth]{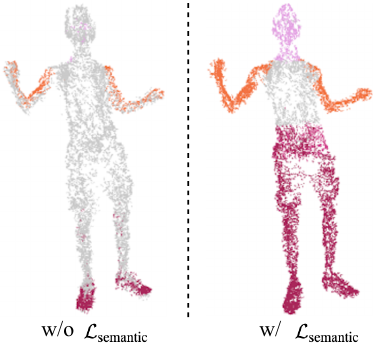}\\
  \caption{\textbf{Ablation Study} on semantic projection with 2D regularization, which enhances semantic accuracy. During pruning, most Gaussians are removed, leaving the remaining ones to default to torso semantics without our semantic supervision.}
  \label{fig:semantic_loss}
\end{figure}

\begin{figure}[ht]
  \centering
    \includegraphics[width=\linewidth]{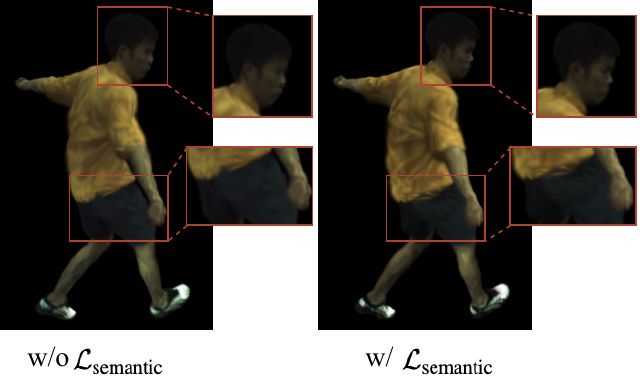}\\
  \caption{\textbf{Ablation Study} on semantic projection with 2D regularization, which keeps the topological consistency of the human body under novel poses.}
  \label{fig:ablation_semantic}
\end{figure}

\subsection{Ablation Study}
\label{sec:ablation}
In this section, we evaluate the effectiveness of our proposed modules through ablation experiments on the ZJU-MoCap~\cite{peng2020neural} dataset. The average metrics over 6 sequences are shown in Table~\ref{tab:ablation}.


\medskip
\noindent
\textbf{Topological and Geometric Feature Learning.} 
As shown in Table~\ref{tab:ablation}, the proposed module significantly (topo-geo) enhances rendering performance. Though it slightly increase inference time, \textit{the notable performance improvement justifies this additional cost}. A qualitative comparison in Fig.~\ref{fig:geo} further proves that Topological and Geometric Feature Learning maintains anatomical coherence during motion and preserving fine details. We also conduct an experiment replacing the sparse 3D U-Net with an MLP (mlp in Table~\ref{tab:ablation}), which demonstrates point-level MLP is limited by a small receptive field, restricting capability to capture the local geometric and topological features.


\medskip
\noindent
\textbf{Semantic Projection with 2D Regularization.} 
This part utilizes semantic labels generated by SHA to supervise the semantic attributes of 3D Gaussians ($\mathcal{L}_{semantic}$). As shown in Fig.~\ref{fig:semantic_loss}, semantic projection with 2D regularization substantially improves the semantic accuracy of 3D Gaussians. At the start of training, Gaussians are neither densified nor pruned~\cite{qian20243dgs}, allowing their scale to grow. During pruning phase, most Gaussians are removed. As a result, most remaining Gaussians default to torso semantics without supervision. The results ($\mathcal{L}_{semantic}$ in Table~\ref{tab:ablation}) highlight the critical role of semantic information. This demonstrates that while the sparse 3D U-Net (introduced in Section~\ref{sec:geometric-semantic}) can capture geometric features with noisy semantic information and improve rendering quality, it still requires accurate semantic data to learn the topology to keep anatomical coherence of the human body, as shown in Fig.~\ref{fig:ablation_semantic}.


\medskip
\noindent
\textbf{Semantic-Guided Density Regularization and Semantic-Aware Regularization with Neighborhood Consistency.} 
Semantic-guided density regularization (sgd) enhances rendering quality by optimizing Gaussian density in areas with high discrepancy, while semantic-aware regularization with neighborhood consistency ($\mathcal{L}_{neighborhood}$) ensures that nearby Gaussians exhibit coherent semantic attributes, thus improving 3D semantic consistency. The improvements in rendering quality are validated by the results in Table~\ref{tab:ablation}.


\section{Conclusion}

In this paper, we propose \our, which uses  semantics-embedded 3D
Gaussians to reconstruct photo-realistic human avatars. \our first integrates a skeleton-driven rigid deformation and a non-rigid cloth dynamics deformation to deform human avatars. \our then leverages SMPL's human body semantic priors to acquire human body semantic labels, which are used to guide optimization of Gaussian’s semantic attribute. We also propose a 3D topology- and geometry-aware network to learn body geometric and topological associations and integrate them into the 3D deformation. We further implement three key strategies to enhance semantic accuracy and render quality: semantic projection with 2D regularization, semantic-guided density regularization, and semantic-aware regularization with neighborhood consistency. Extensive experiments demonstrate that \our outperforms SOTA methods in creating photo-realistic avatars, further validating our hypothesis that integrating semantic priors enhances fine-detail reconstruction. We hope that our method will foster further research in high-quality clothed human avatar synthesis from monocular views.

\medskip
\noindent
\textbf{Limitations.} 
1). \our lacks the capability to extract 3D meshes. Developing a method to extract meshes from 3D Gaussians is an important direction for future research.
2). Topological and Geometric Feature Learning employs a sparse 3D U-Net, which is computationally intensive and may increase training and inference time to some extent.

{\small
\bibliographystyle{ieee_fullname}
\bibliography{egbib}
}

\end{document}